\documentclass[letterpaper,conference]{IEEEtran}

\newif\ifarxiv
\arxivtrue

\IEEEoverridecommandlockouts
\usepackage{cite}
\usepackage{amsmath,amssymb,amsfonts}
\usepackage{algorithmic}
\usepackage{graphicx}
\usepackage{textcomp}
\usepackage{xcolor}
\usepackage{multirow}
\usepackage{makecell}
\newif\ifieee
\ieeetrue

\ifieee
\else
\usepackage[colorlinks=true,allcolors=black]{hyperref} %
\fi

\usepackage[capitalise,noabbrev]{cleveref} %
\usepackage{tikz}
\usetikzlibrary{patterns}
\usepackage{pgfplots}
\pgfplotsset{compat=1.18}
\usetikzlibrary{external}%

\def\BibTeX{{\rm B\kern-.05em{\sc i\kern-.025em b}\kern-.08em
    T\kern-.1667em\lower.7ex\hbox{E}\kern-.125emX}}

\usepackage[caption=false]{subfig}
\usepackage[nolist]{acronym}
\usepackage{balance}

\makeatletter
\newcommand{\linebreakand}{%
  \end{@IEEEauthorhalign}
  \hfill\mbox{}\par
  \mbox{}\hfill\begin{@IEEEauthorhalign}
}
\makeatother

\begin{acronym}[TDMA]
	\acro{AUC-ROC}{Area under the ROC Curve}
    \acro{CNN}{Convolutional Neural Network}
    \acro{EER}{Equal Error Rate}
    \acro{FPR}{False Positive Rate}
    \acro{GMM}{Gaussian Mixture Model}
    \acro{LFW}{Labelled Faces in the Wild}
    \acro{MLP}{Multilayer Perceptron}
    \acro{ROC}{Receiver Operating Characteristic}
\end{acronym}

\usepackage{transparent}
\usepackage{tikz}
\ifarxiv
    \newcommand\copyrighttext{%
      \scriptsize Accepted at SMC 2026. The final published version will be available in IEEE Xplore.}
    \newcommand\copyrightnotice{%
    \begin{tikzpicture}[remember picture,overlay]
    \node[anchor=south,yshift=30pt,xshift=0pt] at (current page.south) {\fbox{\transparent{0.85}\parbox{\dimexpr0.48\textwidth-\fboxsep-\fboxrule\relax}{\copyrighttext}}};
    \end{tikzpicture}%
    }
\else
\fi

\begin{document}

\title{How Far is Too Far? Defining the Distance Threshold for Verification Siamese Networks.
\thanks{This work was funded by the Ministry of Health's R\&D project between SAPS/MS and C3SL/UFPR, by the Brazilian National Council for Scientific and Technological Development (CNPq) – Grant 444192/2024-7, and by a CNPq Research Productivity grant.}}

\author{
   \IEEEauthorblockN{Heloísa Dias Viotto\IEEEauthorrefmark{1}, 
       Cauê Samonek\IEEEauthorrefmark{1},
       Lucas Garcia Pedroso\IEEEauthorrefmark{2},\\
       Marcos Sunye\IEEEauthorrefmark{1},
       André Abed Grégio\IEEEauthorrefmark{1},
       Paulo Lisboa de Almeida\IEEEauthorrefmark{1}}
   \IEEEauthorblockA{\IEEEauthorrefmark{1}
       Departamento de Informática (DInf),
	Universidade Federal do Paran\'a,
	Curitiba, PR - Brazil\\
       \{heloisa.viotto,cauesamonek,sunye,gregio,paulorla\}@ufpr.br
   }
    \IEEEauthorblockA{\IEEEauthorrefmark{2}
        Departamento de Matemática (DMAT),
	 Universidade Federal do Paran\'a
	 Curitiba, PR - Brazil\\
        lucaspedroso@ufpr.br
    }
}

\maketitle

\ifarxiv
    \copyrightnotice
\else
\fi

\begin{abstract}
Siamese verification networks are widely used to compare items such as faces, cars, or signatures. In these scenarios, the network is trained to learn an embedding space in which similar objects are mapped closer together, while dissimilar objects are mapped further apart.
Two objects are considered to belong to the same class (e.g., the same person in two different images) when the distance between their embeddings falls below a predefined threshold. Defining this threshold, however, is a non-trivial task and typically requires labeled data.
In this work, we assume that the distribution of distances produced by a siamese verification network can be approximated by a bimodal function. Based on this assumption, we propose an unsupervised method to determine the verification threshold by identifying the minimum point between the two modes.
The proposed approach does not require annotated samples, enabling the verification threshold to be updated directly in the deployment environment without the cost of manual labeling.
We evaluate our method on four datasets: MNIST, CIFAR-10, LFW, and PKLot. The results indicate that the proposed approach achieves an average verification accuracy of 94\%, comparable to the Equal Error Rate method, while eliminating the need for labeled data.
\end{abstract}


\section{Introduction}

Siamese networks have become ubiquitous due to their effectiveness in tasks such as object tracking, matching, and reidentification~\cite{chicco2021siameseSurvey,ondravsovivc2021siamese,sun2014identificationVerification}. This paper focuses on defining the distance threshold $t$ for verification-by-distance tasks, in which two objects are projected into an embedding space $\mathbb{R}^E$ and compared using a distance metric~\cite{chicco2021siameseSurvey,sun2014identificationVerification}.

\begin{figure}[htbp] \centering \subfloat[General scheme.]{\hspace{-0.05cm}\includegraphics[width=0.22\textwidth]{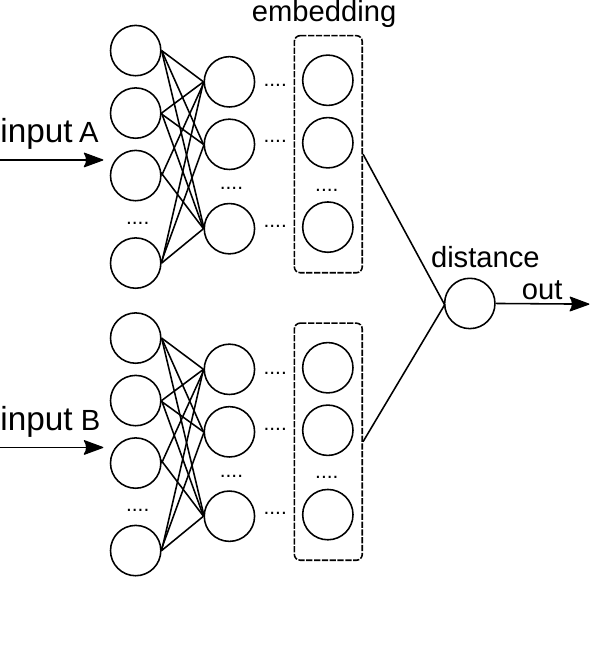}} \subfloat[2D embedding example.]{\hspace{0.05cm}\begin{tikzpicture}[font=\footnotesize]
\begin{axis}[
  xlabel={First Dimension}, ylabel={Second Dimension},
  y label style={at={(-0.20,0.5)}},
  label style={font=\footnotesize},
  xmin=-1, xmax=1, ymin=-1, ymax=1,
  legend style={at={(0.60,0.91)}, anchor=center,legend columns=4, nodes={scale=0.8, transform shape}},
  width=4.7cm, height=4.7cm,
  mark size=2pt,
  only marks
]

\draw (axis cs:0.07,-0.2) -- (axis cs:0.31, -0.17);
\draw[dashed] (axis cs:0.07,-0.2) -- (axis cs:0.69, -0.5);
\draw[dashed] (axis cs:0.07,-0.2) -- (axis cs:-0.52, -0.027);
\draw[dashed] (axis cs:0.07,-0.2) -- (axis cs:0.37, 0.23);

\addplot+[only marks, mark=*, mark options={color=black, fill=lightgray, line width=0pt}] coordinates {
  (-0.5777648737056269, -0.18263214307265752)
  (-0.5232262245697683, -0.027679783921096934)
  (-0.6565724233027538, -0.23233417537221157)
  (-0.6207375975375945, -0.06577355560391873)
  (-0.6705157352413633, -0.08739668388147792)
  (-0.6037147736193711, -0.2624129490033926)
  (-0.5603345461708323, -0.17060777944466188)
  (-0.7246393553014343, -0.02002423613594284)
};
\addlegendentry{\# 0}
\addplot+[only marks, mark=otimes*, mark options={color=black, fill=green, line width=0pt}] coordinates {
  (0.6647151112773193, -0.8443893771868948)
  (0.7530682355493192, -0.7755622634637699)
  (0.7935182373154506, -0.5543140368287102)
  (0.6775604116821279, -0.9070540544218297)
  (0.702610313707704, -0.7968727079304901)
  (0.6929604173544426, -0.6365235414307138)
  (0.5713475853929144, -0.8656024304067584)
  (0.6566373774226157, -0.6279199117650978)
  (0.6609050653442679, -0.7512559160551572)
  (0.7357031443391073, -0.6591340201161727)
  (0.6593677080945861, -0.9118595679490608)
  (0.8147800113079431, -0.7000384506627129)
  (0.5620096193241753, -0.9820864938289073)
  (0.7095210369744154, -0.7053729546858548)
  (0.6711386551280054, -0.7184452538295885)
  (0.6864694265755289, -0.640322013791655)
  (0.6899119478940967, -0.5078988077159801)
  (0.5451279448003803, -0.6863839413731767)
};
\addlegendentry{\# 1}
\addplot+[only marks, mark=square*, mark options={color=black, fill=cyan, line width=0pt}] coordinates {
  (0.31100728694919333, -0.17232786239607123)
  (0.5579793596408023, 0.13731998382179644)
  (0.0728735368684712, -0.20466809711396317)
  (0.6088826945997916, 0.08153530888663418)
  (0.4161545552474768, 0.08282714998658114)
  (0.29395516077414596, -0.10050661125909788)
  (0.5577365695128313, 0.09354960032185411)
  (0.6868784352839434, -0.07208551094408311)
  (0.4540280749484942, 0.08000226949390532)
  (0.5235185886603275, -0.1647197946248724)
  (0.47722577332442007, 0.0731295430398422)
  (0.40536765607484937, 0.0479171346540499)
  (0.5093113682738211, 0.018519663378233853)
  (0.42769733976427715, -0.05158778990905566)
};
\addlegendentry{\# 2}
\addplot+[only marks, mark=triangle*, mark options={color=black, fill=orange, line width=0pt}] coordinates {
  (0.1355848817475207, 0.4946647479094901)
  (0.21265151042316077, 0.624709990311459)
  (0.0915777365121313, 0.4006446844577036)
  (0.3472459947036759, 0.23138315020287936)
  (0.26371857115422803, 0.4758276991392736)
  (0.1436577710878535, 0.5943393101066252)
  (0.23314959140754365, 0.46117822462810176)
  (0.19540174687544098, 0.44473424336797307)
  (0.11748840996970955, 0.5753273560613299)
  (0.16546020229092862, 0.42147697138727813)
};
\addlegendentry{\# 3}
\end{axis}
\end{tikzpicture}} \caption{(a) General scheme of a verification siamese network. (b) Example of a 2D embedding generated by a verification siamese network for the MNIST~\cite{lecun1998MNIST} dataset for the digits 0 to 3. Solid lines represent distances between objects of the same class and dashed lines indicate distances between different classes.} \label{fig:siameseGeneralScheme} \end{figure}
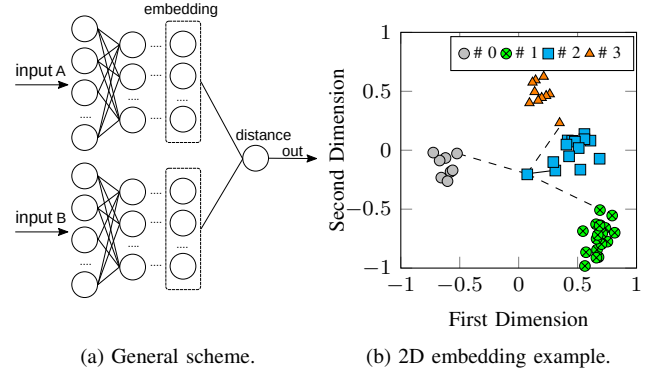

In verification problems~\cite{bromley1993signature,sun2014identificationVerification}, siamese networks are typically composed of two identical branches that share the same weights. Each branch implements a function $f(\mathbf{x}): \mathbb{R}^I \rightarrow \mathbb{R}^E$, responsible for mapping the input space $\mathbb{R}^I$ into the embedding space $\mathbb{R}^E$, where often $I > E$. During training, the network parameters are optimized to project similar inputs close to each other in the embedding space, while mapping dissimilar inputs further apart~\cite{chicco2021siameseSurvey,ondravsovivc2021siamese}. A general scheme of such networks is shown in Figure~\ref{fig:siameseGeneralScheme}a, and an example of a 2D embedding for digit comparison is shown in Figure~\ref{fig:siameseGeneralScheme}b.

After training, two inputs $A$ and $B$ can be compared by feeding them to the network to generate their embeddings. These embeddings are then compared using a distance function $d(f(A), f(B))$, where a common choice is the L2 distance. The inputs $A$ and $B$ are considered to belong to the same class (e.g., two images of the same person) if $d(f(A), f(B)) < t$, where $t$ is a verification threshold~\cite{schroff2015facenet}.

Defining $t$ for siamese verification networks poses several challenges. The maximum distance between embeddings may be unbounded and, although several loss functions aim to enforce a margin $m$ to separate negative pairs (i.e., inputs from different classes)~\cite{schroff2015facenet,hofferAilon2015TripletNet}, using this margin directly as a verification threshold may lead to suboptimal results, since it is difficult to guarantee a margin $m$ for every negative pair during training.

A common practice for setting $t$ is to estimate it using a separate holdout set, or even the training set, for instance by selecting a threshold based on the \ac{ROC} curve~\cite{bromley1993signature,schroff2015facenet,sun2014identificationVerification} (e.g., by considering the \ac{EER}). Despite its popularity, this approach typically requires labeled data and may bias the threshold $t$ toward the distribution of the training or validation set, which may not reflect the distribution encountered during~deployment.

To address this issue, we propose modeling the distances computed over a large number of input pairs as a bimodal function. This assumption is intuitive in verification scenarios, as the first mode is expected to correspond to pairs of objects from the same class, which tend to have smaller distances, while the second mode corresponds to pairs from different classes, with larger distances. By assuming a bimodal behavior, we define the threshold $t$ as the minimum point between the two modes, as exemplified in~\cref{fig:thresholdchoice}. Our approach operates without labeled data, allowing $t$ to be dynamically updated in the deployment environment without the cost of manual data annotation.

The contributions of this work are summarized as follows:

\begin{enumerate}
	\item We propose an unsupervised method for estimating the verification threshold $t$ by modeling embedding distances as a bimodal function.
	\item We show that the proposed approach allows the verification threshold to be updated directly on test data without the need for labeled samples.
\end{enumerate}

The remainder of this paper is organized as follows. Section~\ref{sec:relatedWorks} presents related works. Section~\ref{sec:proposed_method} describes the proposed method. The experimental protocol, including four distinct comparison problems, is detailed in Section~\ref{sec:exp_protocol}. Experimental results are presented in Section~\ref{sec:experiments}, where we evaluate the proposed approach under both balanced and unbalanced scenarios. Finally, Section~\ref{sec:conclusion} concludes the paper.

\section{Related Works}\label{sec:relatedWorks}

Siamese neural networks were first introduced by~\cite{bromley1993signature}, where the authors proposed using two identical \acp{CNN} branches to perform signature verification. Cosine similarity was employed as the distance metric in the comparison head. To detect forgeries, the embedding of a questioned signature was compared against the distribution of embeddings obtained from genuine signatures.

In more recent works, two main training strategies are commonly adopted. The first strategy involves training on pairs of samples $(x_a, x_b)$, where each pair is labeled as positive if $x_b$ belongs to the same class as $x_a$, and negative otherwise~\cite{sun2014identificationVerification,ribas2024DwellTime,pei2023FaceSpoofing,zagoruyko2015ComparePatches}. The training objective minimizes the distance between samples in positive pairs while maximizing it for negative pairs. The second strategy relies on triplets of samples in the form $(x_a, x_p, x_n)$~\cite{schroff2015facenet,balntas2016learning}, where $x_a$ is an anchor sample, $x_p$ is a positive sample from the same class as $x_a$, and $x_n$ is a negative sample from a different class. In this case, the loss function enforces that the embedding of $x_a$ is closer to $x_p$ than to $x_n$ by at least a margin $m$.

In~\cite{sun2014identificationVerification,ribas2024DwellTime}, siamese networks trained using pairs are applied to verification tasks involving faces~\cite{sun2014identificationVerification} and cars~\cite{ribas2024DwellTime}. In both works, the distance $d$ is positively unbounded, i.e., there is no $\Delta > 0$ such that $d(f(A),f(B)) < \Delta$ for the infinitely many possibilities for $A$ and $B$. In~\cite{sun2014identificationVerification}, identity information is leveraged during training to reduce the overlap between embeddings of different individuals. Similarly,~\cite{zagoruyko2015ComparePatches} adopts a pairwise approach using siamese networks, among other architectures, focusing on comparing patches within the same~image.

In~\cite{schroff2015facenet}, the authors employ \acp{CNN} followed by L2 normalization and a triplet loss for face verification. The embeddings are constrained to lie on an $E$-dimensional unit hypersphere, and a hard negative mining strategy is introduced to accelerate convergence by selecting difficult triplets. A related approach is proposed in~\cite{balntas2016learning}, which introduces the in-triplet mining of hard negatives, a lightweight technique also adopted in our experimental protocol (see Section~\ref{sec:exp_protocol}). Triplet-based learning is further explored in~\cite{hofferAilon2015TripletNet}, where each branch of the network processes a different element of the triplet, and in~\cite{wang2014learningTripletNetFineGrain}, which applies a similar strategy to fine-grained classification tasks.

Overall, prior work on siamese networks has addressed challenges related to network architectures~\cite{bromley1993signature,hofferAilon2015TripletNet,wang2014learningTripletNetFineGrain}, the generation of informative training samples~\cite{schroff2015facenet,balntas2016learning}, and applications to specific domains~\cite{bromley1993signature,ribas2024DwellTime}.

Regarding the definition of the comparison threshold $t$, several studies~\cite{bromley1993signature,schroff2015facenet,sun2014identificationVerification,ribas2024DwellTime} rely on labeled training or validation sets, typically by analyzing the \ac{ROC} curve or by selecting thresholds at fixed operating points, such as a predefined \ac{FPR} (e.g., \ac{FPR}95)~\cite{zagoruyko2015ComparePatches}. While effective, these approaches require annotated data and often lead to thresholds that are biased toward the distribution of the dataset used for tuning, which may not generalize well to unseen data in deployment scenarios.

In contrast, the method proposed in this work estimates the threshold $t$ directly from the distribution of pairwise distances, enabling it to be updated dynamically, even during testing, without the need for labeled samples.

\section{Proposed Method}\label{sec:proposed_method}

In this work, we hypothesize that the distances between objects computed by a siamese network follow a distribution that can be reasonably approximated by a bimodal one. This distribution can then be exploited to estimate a verification threshold, enabling the classification of object pairs as belonging to either the same or different classes.

Under the bimodal assumption, the verification threshold can be estimated without requiring labeled data, allowing the verification threshold to be updated directly in the deployment environment. As new data is processed, the underlying distance distribution can be recomputed, enabling the threshold to adapt over time. This property is particularly relevant in scenarios subject to virtual concept drift, where the data distribution evolves while class definitions remain unchanged~\cite{barboza2025incades}.

To estimate the threshold, we first compute distances between multiple pairs of objects. To obtain these distances, it is not necessary to know whether each pair is positive or negative (i.e., whether the objects belong to the same class). Nevertheless, it is desirable that the number of positive and negative pairs be relatively balanced (although we show an experiment where the proposed method is able to cope with an unbalanced scenario in Section~\ref{sec:experiments}). Such pairs can be generated, for instance, from a validation set or even from unlabeled data collected in the target deployment environment.

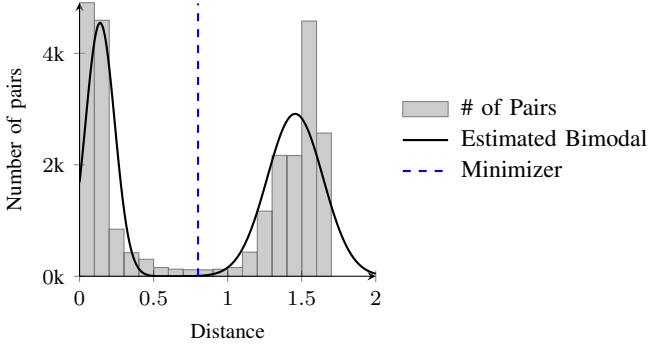
\begin{figure}[htbp]
	\centering
	\begin{tikzpicture}
    \begin{axis}[
    	ybar,
    	width=5.5cm,
    	height=5.2cm,
    	scaled y ticks=base 10:-3,
    	ytick scale label code/.code={},
    	yticklabel={\pgfmathprintnumber{\tick}k},
    	ticklabel style = {font=\footnotesize},
    	label style={font=\footnotesize},
    	axis x line=bottom,
    	axis y line=left,
    	xlabel=Distance,
    	ylabel=Number of pairs,
    	xmin=0, xmax=2,
    	y label style={at={(-0.15,0.5)}},
    	legend style={at={(0.4,1.05)},
    		anchor=south,    
    		legend columns=2, %
    		/tikz/every even column/.append style={column sep=0.2cm}, %
    		nodes={scale=0.8, transform shape}
    	},
    ]

    \addplot+ [ybar interval, opacity=0.5, color=black, mark=none, fill=gray!80, draw=black]
        table [row sep=\\,x=Bin,y=Equal]{
            Bin Equal \\
            0.0 4907 \\
			0.1 4592 \\
			0.2 844 \\
			0.3 422 \\
			0.4 304 \\
			0.5 156 \\
			0.6 127 \\
			0.7 117 \\
			0.8 114 \\
			0.9 126 \\
			1.0 156 \\
			1.1 433 \\
			1.2 1168 \\
			1.3 2169 \\
			1.4 2166 \\
			1.5 4578 \\
			1.6 2569 \\
			1.7 52 \\
        };

    \addplot [const plot=false, black, thick, mark=none, domain=0:2, samples=100, smooth]
    {2500.00000 * (
        0.55078 * (1/(sqrt(2*pi*0.18859^2)) * exp(-((x - 1.45571)^2)/(2*0.18859^2)))
      + 0.44922 * (1/(sqrt(2*pi*0.09854^2)) * exp(-((x - 0.13882)^2)/(2*0.09854^2)))
    )};
    
    \addplot+ [sharp plot, color=blue, thick, mark=none, dashed] coordinates {(0.8, 0) (0.8, 4893)};

    \end{axis}
\end{tikzpicture} \hspace{-0.2cm}\raisebox{1.5\height}{\begin{tikzpicture}

\begin{axis}[%
hide axis,
xmin=-0.48,
xmax=7.5,
ymin=3.0,
ymax=9.0,
ylabel={$f_2$},
axis x line*=bottom,
axis y line*=left,
legend style={legend cell align=left,align=left,draw=black,legend columns=1, column sep=2.5pt, draw=none, font=\small}
]

\addlegendimage{%
	area legend,
	draw=black,
	fill=gray!80,
	opacity=0.5,
}
\addlegendentry{\# of Pairs};

\addlegendimage{color=black,solid,thick}
\addlegendentry{Estimated Bimodal};

\addlegendimage{color=blue,dashed,thick}
\addlegendentry{Minimizer};

\end{axis}
\end{tikzpicture}%
}
	\caption{Fitted bimodal curve and verification threshold selection.}
	\label{fig:thresholdchoice}
\end{figure}

We use the resulting distances to model a distribution and subsequently apply an optimization algorithm to infer the optimal classification threshold. An observational analysis reveals that the distance distribution often exhibits two distinct peaks. We decided to fit a curve to this distribution and set the classification threshold $t$ as the local minimizer between these peaks, as exemplified in Figure~\ref{fig:thresholdchoice}. Since the distribution resembles a mixture of two normal distributions, we chose to fit a \ac{GMM} with two components, defining $t$ as the minimizer located between the two modes.

Note that by modeling the problem as a bimodal distribution, we can guarantee the existence of a global minimum strictly inside the interval between the two modes.

\section{Experimental Protocol}\label{sec:exp_protocol}

\subsection{Datasets and Triplets Generation}

We evaluate the proposed approach on four well-known datasets: MNIST~\cite{lecun1998MNIST}, \ac{LFW}~\cite{huang2008LFW}, PKLot~\cite{almeida2015PKLot}, and CIFAR-10~\cite{krizhevsky2009Cifar10}. These datasets allow us to assess the method across distinct verification tasks, including digit comparison (MNIST), vehicle verification (PKLot), face verification (LFW), and generic object category comparison (CIFAR-10). For MNIST, CIFAR-10, and \ac{LFW}, we adopt the training and testing splits commonly used in the literature~\cite{lecun1998MNIST,krizhevsky2009Cifar10,huang2008LFW,hofferAilon2015TripletNet}. For PKLot, we follow the cross-camera evaluation protocol proposed in~\cite{almeida2015PKLot,ribas2024DwellTime}, resulting in three distinct experimental configurations. As in~\cite{ribas2024DwellTime}, the task in PKLot is to verify whether two images of vehicles, captured at different times in the same parking spot, correspond to the same car.

Training triplets are generated as follows. Each image in the training set is used as an anchor $x_a$, paired with another image from the same class as the positive sample $x_p$. The negative sample $x_n$ is randomly selected from a different class. This procedure results in one triplet per training sample at each epoch (e.g., 60,000 triplets per epoch for MNIST).

During testing, verification is performed using image pairs. For MNIST and CIFAR-10, we generate 10,000 balanced test pairs, with 50\% positive and 50\% negative pairs. Each test image is used as an anchor and randomly paired with either an image from the same class or from a different class, with equal probability. The same procedure is adopted for \ac{LFW}, yielding 1,642 test pairs. For PKLot, we follow a protocol similar to~\cite{ribas2024DwellTime}. For each vehicle in the test set, one image of the same car captured at a different time is selected to form a positive pair, and one image of a different vehicle is selected to form a negative pair. This results in 17,436 test pairs for PUCPR, 5,976 for UFPR05, and 3,806 for UFPR04.

The number of training triplets and testing pairs for each dataset is summarized in Table~\ref{table:datasets}. In all cases, the verification task consists of deciding whether the two images in each pair belong to the same class.

\begin{table}[htpb] \centering \caption{Datasets used in the experiments.} \label{table:datasets} \begin{tabular}{lrr} \hline Dataset & \# Training Triplets & \# Testing Pairs\\\hline MNIST & 60,000 & 10,000 \\ CIFAR-10 & 50,000 & 10,000 \\ LFW & 8,343 & 1,642 \\ PKLot 1 & 170,924 (UFPR04 + UFPR05) & 17,436 (PUCPR) \\ PKLot 2 & 433,562 (UFPR04 + PUCPR) & 5,976 (UFPR05) \\ PKLot 3 & 480,750 (UFPR05 + PUCPR) & 3,806 (UFPR04)\\\hline \end{tabular} \end{table}

\subsection{Networks and Training Procedure}

Following~\cite{ribas2024DwellTime}, we employ MobileNetV3-Large~\cite{howard2019MobileNetV3}, initialized with ImageNet weights, as the backbone of all siamese networks. The classification layers are removed, and an L2 normalization layer is appended after the final convolutional layer, as in~\cite{schroff2015facenet}, constraining the embeddings to lie on an $e$-dimensional unit hypersphere. The output of this layer is used as the embedding representation.

Input images are resized to $128 \times 128$ pixels with three channels\footnote{For grayscale datasets, the single channel is replicated three times for consistency across experiments.}, producing embeddings of dimension 960. Training is performed using the Adam optimizer with an initial learning rate of 0.001 and a batch size of 64. We adopt the triplet loss proposed in~\cite{wang2014learningTripletNetFineGrain}, defined as
\[
L = \max(0, m + \|f(x_a) - f(x_p)\|_2 - \|f(x_a) - f(x_n)\|_2),
\]
where $m$ denotes the margin, fixed to 1 in all experiments. Networks are trained for 30 epochs. Each training set is randomly split into 70\% for training and 30\% for validation. The validation set is used exclusively for threshold estimation, both for the \ac{EER} baseline and for the proposed method. Each validation triplet is decomposed into two pairs: a positive pair $(x_a, x_p)$ and a negative pair $(x_a, x_n)$ in the validation set.

\subsection{Tested Methods}

We evaluate the proposed approach under two configurations. The first, referred to as \textit{BimodalVal}, estimates the verification threshold using the validation set only. The second, called \textit{BimodalUpd}, initializes the threshold using the validation set and subsequently updates both the distance distribution and the threshold $t$ during testing. In this setting, the threshold is recomputed every 64 newly observed distances, using a sliding window containing the 1,024 most recent samples.

For the proposed method, we use the \texttt{sklearn.mixture.GaussianMixture} implementation, which fits a Gaussian mixture model via the expectation-maximization algorithm~\cite{dempster1977ExpectationMaximization}. All parameters are kept at their default values, except for the number of mixture components, which is set to two. The threshold is obtained by minimizing the fitted bimodal density using \texttt{scipy.optimize.minimize\_scalar}, with bounds defined by the means of the two Gaussian components. This procedure employs Brent’s derivative-free optimization algorithm~\cite{brent2013algorithms}. During training, we also apply in-triplet hard negative mining with anchor swapping, as proposed in~\cite{balntas2016learning}.

As baselines, we evaluate a method based on the \ac{ROC} curve, where the threshold is selected as the point corresponding to the \ac{EER} on the validation set. Additionally, we report results obtained by directly using the margin $m$ as the verification threshold $t$.

\section{Experiments and Results}\label{sec:experiments}

We first present results obtained under balanced test scenarios in Section~\ref{sec:normalizedTest}. 
We then analyze the impact of removing embedding normalization, resulting in positively unbounded distances, in Section~\ref{sec:non_normalizedTests}. 
Finally, we evaluate the proposed approach under unbalanced testing conditions in Section~\ref{sec:unbalancedTest}. 
All reported results are averaged over five independent runs.

\subsection{Normalized Embeddings}\label{sec:normalizedTest}

Table~\ref{table:results} reports the verification accuracies obtained following the experimental protocol described in Section~\ref{sec:exp_protocol}. 
The average threshold values estimated by the proposed method and by the \ac{EER} criterion are shown in Table~\ref{table:thresholds}. 
Since the threshold is updated during testing for the \textit{BimodalUpd} configuration, we report its average value. 
Because the test sets are balanced between positive and negative pairs, accuracy is used as the evaluation metric.

\begin{table}[htpb] \caption{Accuracies achieved considering the tested methods.} \label{table:results} \centering \begin{tabular}{lrrrr}\hline & & &\multicolumn{2}{c}{Proposed Methods} \\ & EER & Margin & BimodalVal & BimodalUpd \\\hline MNIST & \textbf{0.99} $\pm$0.00 & \textbf{0.99} $\pm$0.00 & \textbf{0.99} $\pm$0.00 & \textbf{0.99} $\pm$0.00 \\ CIFAR-10 & \textbf{0.90} $\pm$0.01 & \textbf{0.90} $\pm$0.01 & 0.87 $\pm$0.01 & 0.88 $\pm$0.01 \\ LFW & \textbf{0.86} $\pm$0.00 & 0.84 $\pm$0.00 & \textbf{0.86} $\pm$0.00 & \textbf{0.86} $\pm$0.00 \\ PKLot 1 & 0.96 $\pm$0.01 & 0.93 $\pm$0.01 & 0.92 $\pm$0.02 & \textbf{0.97} $\pm$0.01 \\ PKLot 2 & 0.95 $\pm$0.02 & \textbf{0.96} $\pm$0.01 & 0.94 $\pm$0.01 & \textbf{0.96} $\pm$0.00 \\ PKLot 3 & \textbf{0.98} $\pm$0.01 & \textbf{0.98} $\pm$0.00 & 0.97 $\pm$0.00 & \textbf{0.98} $\pm$0.00 \\\hline Average & \textbf{0.94} & 0.93 & 0.92 & \textbf{0.94} \\ \hline \end{tabular} \end{table} \begin{table}[htpb] \caption{Thresholds computed using our proposed method versus the approach that considers the \ac{EER}.} \label{table:thresholds} \centering \begin{tabular}{lrrr}\hline &&\multicolumn{2}{c}{Proposed Methods}\\ & EER & BimodalVal & BimodalUpd \\\hline MNIST & 1.14 $\pm$0.02 & 0.43 $\pm$0.03 & 0.45 $\pm$0.03 \\ CIFAR-10 & 1.11 $\pm$0.03 & 0.45 $\pm$0.08 & 0.69 $\pm$0.06 \\ LFW & 1.20 $\pm$0.02 & 1.28 $\pm$0.01 & 1.28 $\pm$0.01 \\ PKLot 1 & 0.85 $\pm$0.07 & 1.04 $\pm$0.01 & 0.73 $\pm$0.03 \\ PKLot 2 & 1.05 $\pm$0.08 & 1.12 $\pm$0.02 & 0.89 $\pm$0.06 \\ PKLot 3 & 0.97 $\pm$0.12 & 0.78 $\pm$0.10 & 1.02 $\pm$0.05 \\\hline \end{tabular} \end{table}

As shown in Table~\ref{table:results}, all methods achieve near-perfect performance on the MNIST, which is a simple dataset. 
This behavior is further illustrated in Table~\ref{table:thresholds}, where substantially different thresholds lead to similar accuracies, indicating a large separation between classes in the embedding space.

For the \ac{LFW} and PKLot~1, directly using the margin as a verification threshold leads to inferior results compared to both the proposed approach and the \ac{EER}-based method. 
This observation reinforces the difficulty of enforcing a suitable margin $m$ for all negative pairs during training under some scenarios, given the combinatorial number of possible triplets.

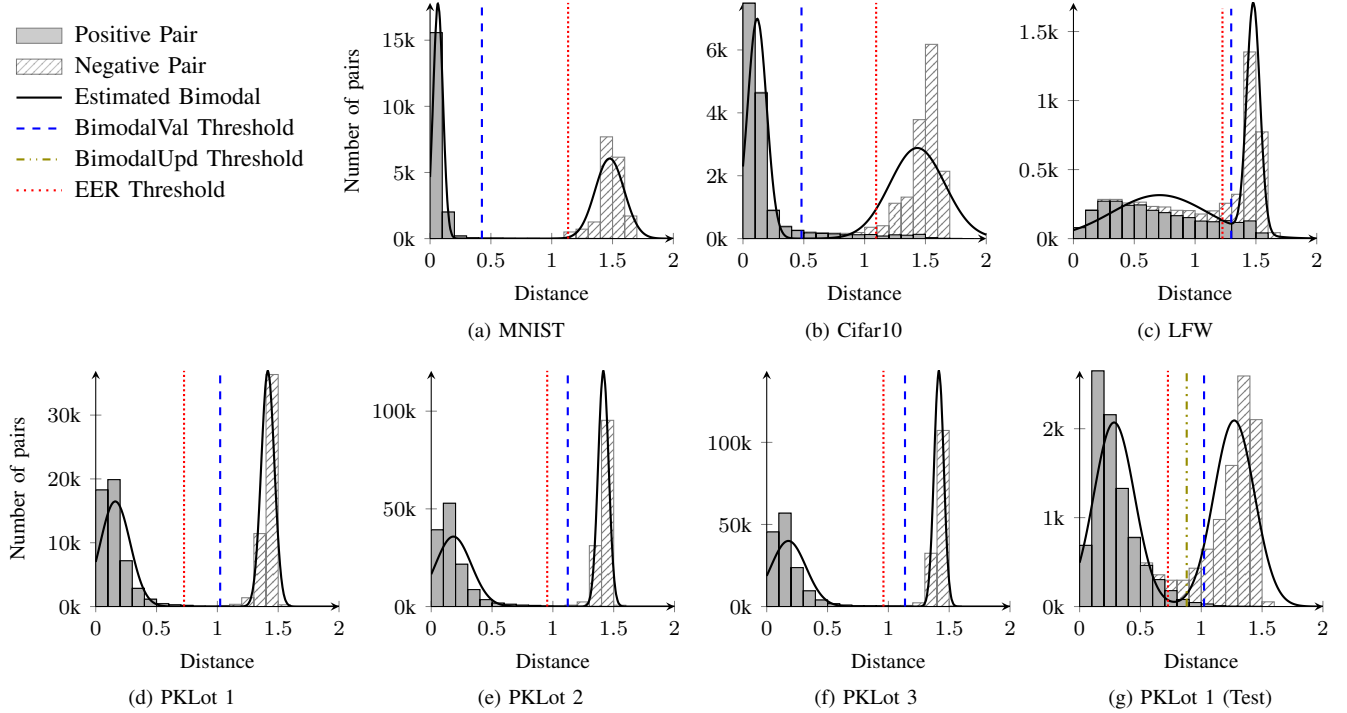
\begin{figure*}[htbp] \centering \raisebox{.5\height}{\begin{tikzpicture}

\begin{axis}[%
hide axis,
xmin=-0.48,
xmax=7.5,
ymin=3.0,
ymax=9.0,
ylabel={$f_2$},
axis x line*=bottom,
axis y line*=left,
legend style={legend cell align=left,align=left,draw=black,legend columns=1, column sep=2.5pt, draw=none,font=\small}
]

\addlegendimage{%
	area legend,
	draw=black,
	fill=gray!80,
	opacity=0.5,
}
\addlegendentry{Positive Pair};

\addlegendimage{%
	area legend,
	draw=black,
	fill=black,
	pattern=north east lines,
	pattern color=black,
	opacity=0.5,
}
\addlegendentry{Negative Pair};

\addlegendimage{color=black,solid,thick}
\addlegendentry{Estimated Bimodal};

\addlegendimage{color=blue,dashed,thick}
\addlegendentry{BimodalVal Threshold};

\addlegendimage{color=olive,dashdotdotted,thick}
\addlegendentry{BimodalUpd Threshold};

\addlegendimage{color=red,dotted,thick}
\addlegendentry{EER Threshold};

\end{axis}
\end{tikzpicture}%
 } \subfloat[MNIST]{\begin{tikzpicture}
    \begin{axis}[
        ybar,                                                                       
        width=4.8cm,                                                                
        height=4.7cm,                                                               
        scaled y ticks=base 10:-3,                                                  
        ytick scale label code/.code={},                                            
        yticklabel={\pgfmathprintnumber{\tick}k},                                   
        ticklabel style = {font=\footnotesize},                                     
        label style={font=\footnotesize},                                           
        axis x line=bottom,                                                         
        axis y line=left,           
        xlabel=Distance,
        ylabel=Number of pairs,
        xmin=0, xmax=2
    ]

    \addplot+ [ybar interval, opacity=1, color=gray, pattern = north east lines, pattern color = gray, mark=none]
        table [row sep=\\,x=Bin,y=Diff]{
            Bin Diff \\
            0.0 15573 \\
			0.1 2010 \\
			0.2 202 \\
			0.3 56 \\
			0.4 25 \\
			0.5 29 \\
			0.6 15 \\
			0.7 11 \\
			0.8 16 \\
			0.9 7 \\
			1.0 26 \\
			1.1 501 \\
			1.2 713 \\
			1.3 1257 \\
			1.4 7693 \\
			1.5 6155 \\
			1.6 1709 \\
			1.7 1 \\
			1.8 0 \\
			1.9 0 \\
        };

    \addplot+ [ybar interval, opacity=1, color=black, mark=none, fill=gray!60, draw=black]
        table [row sep=\\,x=Bin,y=Equal]{
            Bin Equal \\
            0.0 15572 \\
			0.1 2008 \\
			0.2 201 \\
			0.3 55 \\
			0.4 24 \\
			0.5 28 \\
			0.6 15 \\
			0.7 11 \\
			0.8 13 \\
			0.9 5 \\
			1.0 13 \\
			1.1 9 \\
			1.2 15 \\
			1.3 12 \\
			1.4 16 \\
			1.5 3 \\
			1.6 0 \\
			1.7 0 \\
			1.8 0 \\
			1.9 0 \\
        };

    \addplot [const plot=false, black, thick, mark=none, domain=0:2, samples=100, smooth]
    {3599.90000 * (
        0.49581 * (1/(sqrt(2*pi*0.03995^2)) * exp(-((x - 0.06547)^2)/(2*0.03995^2)))
      + 0.50419 * (1/(sqrt(2*pi*0.11960^2)) * exp(-((x - 1.47618)^2)/(2*0.11960^2)))
    )};
    
    \addplot+ [ sharp plot, color=blue, thick, mark=none, dashed] coordinates {(0.4261264962721833, 0) (0.4261264962721833, 17822.5793430079)};

    \addplot+ [ sharp plot, color=red, thick, mark=none, densely dotted] coordinates {(1.1347757577896118, 0) (1.1347757577896118, 17822.5793430079)};

    \end{axis}
\end{tikzpicture}} \subfloat[Cifar10]{\begin{tikzpicture}
    \begin{axis}[
        ybar,                                                                       
        width=4.8cm,                                                                
        height=4.7cm,                                                               
        scaled y ticks=base 10:-3,                                                  
        ytick scale label code/.code={},                                            
        yticklabel={\pgfmathprintnumber{\tick}k},                                   
        ticklabel style = {font=\footnotesize},                                     
        label style={font=\footnotesize},                                           
        axis x line=bottom,                                                         
        axis y line=left,           
        xlabel=Distance,
        xmin=0, xmax=2
    ]

    \addplot+ [ybar interval, opacity=1, color=gray, pattern = north east lines, pattern color = gray, mark=none]
        table [row sep=\\,x=Bin,y=Diff]{
            Bin Diff \\
            0.0 7497 \\
			0.1 4653 \\
			0.2 915 \\
			0.3 397 \\
			0.4 273 \\
			0.5 211 \\
			0.6 191 \\
			0.7 175 \\
			0.8 168 \\
			0.9 194 \\
			1.0 360 \\
			1.1 409 \\
			1.2 1124 \\
			1.3 1319 \\
			1.4 3783 \\
			1.5 6178 \\
			1.6 2141 \\
			1.7 11 \\
			1.8 0 \\
			1.9 0 \\
        };

    \addplot+ [ybar interval, opacity=1, color=black, mark=none, fill=gray!60, draw=black]
        table [row sep=\\,x=Bin,y=Equal]{
            Bin Equal \\
            0.0 7485 \\
			0.1 4631 \\
			0.2 891 \\
			0.3 377 \\
			0.4 256 \\
			0.5 183 \\
			0.6 164 \\
			0.7 151 \\
			0.8 123 \\
			0.9 123 \\
			1.0 129 \\
			1.1 82 \\
			1.2 120 \\
			1.3 108 \\
			1.4 135 \\
			1.5 31 \\
			1.6 11 \\
			1.7 0 \\
			1.8 0 \\
			1.9 0 \\
        };

    \addplot [const plot=false, black, thick, mark=none, domain=0:2, samples=100, smooth]
    {2999.90000 * (
        0.45580 * (1/(sqrt(2*pi*0.07799^2)) * exp(-((x - 0.11649)^2)/(2*0.07799^2)))
      + 0.54420 * (1/(sqrt(2*pi*0.22603^2)) * exp(-((x - 1.43312)^2)/(2*0.22603^2)))
    )};
    
    \addplot+ [ sharp plot, color=blue, thick, mark=none, dashed] coordinates {(0.4810590471675446, 0) (0.4810590471675446, 7497)};

    \addplot+ [ sharp plot, color=red, thick, mark=none, densely dotted] coordinates {(1.0946788787841797, 0) (1.0946788787841797, 7497)};

    \end{axis}
\end{tikzpicture}} \subfloat[LFW]{\begin{tikzpicture}
    \begin{axis}[
        ybar,                                                                       
        width=4.8cm,                                                                
        height=4.7cm,                                                               
        scaled y ticks=base 10:-3,                                                  
        ytick scale label code/.code={},                                            
        yticklabel={\pgfmathprintnumber{\tick}k},                                   
        ticklabel style = {font=\footnotesize},                                     
        label style={font=\footnotesize},                                           
        axis x line=bottom,                                                         
        axis y line=left,           
        xlabel=Distance,
        xmin=0, xmax=2
    ]

    \addplot+ [ybar interval, opacity=1, color=gray, pattern = north east lines, pattern color = gray, mark=none]
        table [row sep=\\,x=Bin,y=Diff]{
            Bin Diff \\
            0.0 80 \\
			0.1 209 \\
			0.2 284 \\
			0.3 284 \\
			0.4 258 \\
			0.5 264 \\
			0.6 233 \\
			0.7 229 \\
			0.8 205 \\
			0.9 203 \\
			1.0 178 \\
			1.1 201 \\
			1.2 256 \\
			1.3 321 \\
			1.4 1352 \\
			1.5 773 \\
			1.6 43 \\
			1.7 0 \\
			1.8 0 \\
			1.9 0 \\
        };

    \addplot+ [ybar interval, opacity=1, color=black, mark=none, fill=gray!60, draw=black]
        table [row sep=\\,x=Bin,y=Equal]{
            Bin Equal \\
            0.0 79 \\
			0.1 204 \\
			0.2 271 \\
			0.3 271 \\
			0.4 241 \\
			0.5 244 \\
			0.6 205 \\
			0.7 188 \\
			0.8 166 \\
			0.9 152 \\
			1.0 127 \\
			1.1 122 \\
			1.2 130 \\
			1.3 117 \\
			1.4 128 \\
			1.5 41 \\
			1.6 1 \\
			1.7 0 \\
			1.8 0 \\
			1.9 0 \\
        };

    \addplot [const plot=false, black, thick, mark=none, domain=0:2, samples=100, smooth]
    {537.30000 * (
        0.42367 * (1/(sqrt(2*pi*0.05454^2)) * exp(-((x - 1.47877)^2)/(2*0.05454^2)))
      + 0.57633 * (1/(sqrt(2*pi*0.39235^2)) * exp(-((x - 0.70880)^2)/(2*0.39235^2)))
    )};
    
    \addplot+ [ sharp plot, color=blue, thick, mark=none, dashed] coordinates {(1.296862913399386, 0) (1.296862913399386, 1664.9751203780936)};

    \addplot+ [ sharp plot, color=red, thick, mark=none, densely dotted] coordinates {(1.2249490022659302, 0) (1.2249490022659302, 1664.9751203780936)};

    \end{axis}
\end{tikzpicture}}\\ \subfloat[PKLot 1]{\begin{tikzpicture}
    \begin{axis}[
        ybar,                                                                       
        width=4.8cm,                                                                
        height=4.7cm,                                                               
        scaled y ticks=base 10:-3,                                                  
        ytick scale label code/.code={},                                            
        yticklabel={\pgfmathprintnumber{\tick}k},                                   
        ticklabel style = {font=\footnotesize},                                     
        label style={font=\footnotesize},                                           
        axis x line=bottom,                                                         
        axis y line=left,           
        xlabel=Distance,
        ylabel=Number of pairs,
        xmin=0, xmax=2
    ]

    \addplot+ [ybar interval, opacity=1, color=gray, pattern = north east lines, pattern color = gray, mark=none]
        table [row sep=\\,x=Bin,y=Diff]{
            Bin Diff \\
            0.0 18276 \\
			0.1 19895 \\
			0.2 7207 \\
			0.3 2870 \\
			0.4 1177 \\
			0.5 479 \\
			0.6 311 \\
			0.7 220 \\
			0.8 87 \\
			0.9 82 \\
			1.0 164 \\
			1.1 351 \\
			1.2 1381 \\
			1.3 11431 \\
			1.4 36360 \\
			1.5 268 \\
			1.6 0 \\
			1.7 0 \\
			1.8 0 \\
			1.9 0 \\
        };

    \addplot+ [ybar interval, opacity=1, color=black, mark=none, fill=gray!60, draw=black]
        table [row sep=\\,x=Bin,y=Equal]{
            Bin Equal \\
            0.0 18274 \\
			0.1 19885 \\
			0.2 7171 \\
			0.3 2844 \\
			0.4 1143 \\
			0.5 445 \\
			0.6 268 \\
			0.7 167 \\
			0.8 39 \\
			0.9 14 \\
			1.0 14 \\
			1.1 5 \\
			1.2 10 \\
			1.3 1 \\
			1.4 0 \\
			1.5 0 \\
			1.6 0 \\
			1.7 0 \\
			1.8 0 \\
			1.9 0 \\
        };

    \addplot [const plot=false, black, thick, mark=none, domain=0:2, samples=100, smooth]
    {10055.90000 * (
        0.50350 * (1/(sqrt(2*pi*0.12262^2)) * exp(-((x - 0.16060)^2)/(2*0.12262^2)))
      + 0.49650 * (1/(sqrt(2*pi*0.05393^2)) * exp(-((x - 1.41472)^2)/(2*0.05393^2)))
    )};
    
    \addplot+ [ sharp plot, color=blue, thick, mark=none, dashed] coordinates {(1.0229352433220382, 0) (1.0229352433220382, 36935.15388846727)};

    \addplot+ [ sharp plot, color=red, thick, mark=none, densely dotted] coordinates {(0.7259594202041626, 0) (0.7259594202041626, 36935.15388846727)};

    \end{axis}
\end{tikzpicture}} \subfloat[PKLot 2]{\begin{tikzpicture}
    \begin{axis}[
        ybar,                                                                       
        width=4.8cm,                                                                
        height=4.7cm,                                                               
        scaled y ticks=base 10:-3,                                                  
        ytick scale label code/.code={},                                            
        yticklabel={\pgfmathprintnumber{\tick}k},                                   
        ticklabel style = {font=\footnotesize},                                     
        label style={font=\footnotesize},                                           
        axis x line=bottom,                                                         
        axis y line=left,           
        xlabel=Distance,
        xmin=0, xmax=2
    ]

    \addplot+ [ybar interval, opacity=1, color=gray, pattern = north east lines, pattern color = gray, mark=none]
        table [row sep=\\,x=Bin,y=Diff]{
            Bin Diff \\
            0.0 39261 \\
			0.1 52873 \\
			0.2 21683 \\
			0.3 8712 \\
			0.4 3570 \\
			0.5 1801 \\
			0.6 1207 \\
			0.7 763 \\
			0.8 641 \\
			0.9 359 \\
			1.0 356 \\
			1.1 731 \\
			1.2 2363 \\
			1.3 31155 \\
			1.4 95318 \\
			1.5 677 \\
			1.6 1 \\
			1.7 0 \\
			1.8 0 \\
			1.9 0 \\
        };

    \addplot+ [ybar interval, opacity=1, color=black, mark=none, fill=gray!60, draw=black]
        table [row sep=\\,x=Bin,y=Equal]{
            Bin Equal \\
            0.0 39261 \\
			0.1 52857 \\
			0.2 21658 \\
			0.3 8673 \\
			0.4 3525 \\
			0.5 1752 \\
			0.6 1156 \\
			0.7 697 \\
			0.8 547 \\
			0.9 219 \\
			1.0 74 \\
			1.1 61 \\
			1.2 63 \\
			1.3 111 \\
			1.4 82 \\
			1.5 0 \\
			1.6 0 \\
			1.7 0 \\
			1.8 0 \\
			1.9 0 \\
        };

    \addplot [const plot=false, black, thick, mark=none, domain=0:2, samples=100, smooth]
    {26147.10000 * (
        0.49765 * (1/(sqrt(2*pi*0.04303^2)) * exp(-((x - 1.41523)^2)/(2*0.04303^2)))
      + 0.50235 * (1/(sqrt(2*pi*0.14632^2)) * exp(-((x - 0.18215)^2)/(2*0.14632^2)))
    )};
    
    \addplot+ [ sharp plot, color=blue, thick, mark=none, dashed] coordinates {(1.1224910701499802, 0) (1.1224910701499802, 120648.13987805127)};

    \addplot+ [ sharp plot, color=red, thick, mark=none, densely dotted] coordinates {(0.9534788727760315, 0) (0.9534788727760315, 120648.13987805127)};

    \end{axis}
\end{tikzpicture}} \subfloat[PKLot 3]{\begin{tikzpicture}
    \begin{axis}[
        ybar,                                                                       
        width=4.8cm,                                                                
        height=4.7cm,                                                               
        scaled y ticks=base 10:-3,                                                  
        ytick scale label code/.code={},                                            
        yticklabel={\pgfmathprintnumber{\tick}k},                                   
        ticklabel style = {font=\footnotesize},                                     
        label style={font=\footnotesize},                                           
        axis x line=bottom,                                                         
        axis y line=left,           
        xlabel=Distance,
        xmin=0, xmax=2
    ]

    \addplot+ [ybar interval, opacity=1, color=gray, pattern = north east lines, pattern color = gray, mark=none]
        table [row sep=\\,x=Bin,y=Diff]{
            Bin Diff \\
            0.0 45450 \\
			0.1 56912 \\
			0.2 23739 \\
			0.3 9617 \\
			0.4 4042 \\
			0.5 1787 \\
			0.6 924 \\
			0.7 524 \\
			0.8 484 \\
			0.9 489 \\
			1.0 486 \\
			1.1 679 \\
			1.2 2276 \\
			1.3 32450 \\
			1.4 107256 \\
			1.5 403 \\
			1.6 3 \\
			1.7 0 \\
			1.8 0 \\
			1.9 0 \\
        };

    \addplot+ [ybar interval, opacity=1, color=black, mark=none, fill=gray!60, draw=black]
        table [row sep=\\,x=Bin,y=Equal]{
            Bin Equal \\
            0.0 45450 \\
			0.1 56882 \\
			0.2 23708 \\
			0.3 9570 \\
			0.4 3994 \\
			0.5 1736 \\
			0.6 867 \\
			0.7 454 \\
			0.8 390 \\
			0.9 341 \\
			1.0 210 \\
			1.1 60 \\
			1.2 20 \\
			1.3 32 \\
			1.4 47 \\
			1.5 0 \\
			1.6 0 \\
			1.7 0 \\
			1.8 0 \\
			1.9 0 \\
        };

    \addplot [const plot=false, black, thick, mark=none, domain=0:2, samples=100, smooth]
    {28752.10000 * (
        0.49685 * (1/(sqrt(2*pi*0.03969^2)) * exp(-((x - 1.41559)^2)/(2*0.03969^2)))
      + 0.50315 * (1/(sqrt(2*pi*0.14439^2)) * exp(-((x - 0.17814)^2)/(2*0.14439^2)))
    )};
    
    \addplot+ [ sharp plot, color=blue, thick, mark=none, dashed] coordinates {(1.1367887276935924, 0) (1.1367887276935924, 143580.62748264687)};

    \addplot+ [ sharp plot, color=red, thick, mark=none, densely dotted] coordinates {(0.9595876932144165, 0) (0.9595876932144165, 143580.62748264687)};

    \end{axis}
\end{tikzpicture}} \subfloat[PKLot 1 (Test)]{\begin{tikzpicture}
    \begin{axis}[
        ybar,                                                                       
        width=4.8cm,                                                                
        height=4.7cm,                                                               
        scaled y ticks=base 10:-3,                                                  
        ytick scale label code/.code={},                                            
        yticklabel={\pgfmathprintnumber{\tick}k},                                   
        ticklabel style = {font=\footnotesize},                                     
        label style={font=\footnotesize},                                           
        axis x line=bottom,                                                         
        axis y line=left,           
        xlabel=Distance,
        xmin=0, xmax=2
    ]

    \addplot+ [ybar interval, opacity=1, color=gray, pattern = north east lines, pattern color = gray, mark=none]
        table [row sep=\\,x=Bin,y=Diff]{
            Bin Diff \\
            0.0 688 \\
			0.1 2651 \\
			0.2 2157 \\
			0.3 1329 \\
			0.4 781 \\
			0.5 491 \\
			0.6 357 \\
			0.7 296 \\
			0.8 296 \\
			0.9 431 \\
			1.0 645 \\
			1.1 979 \\
			1.2 1587 \\
			1.3 2593 \\
			1.4 2102 \\
			1.5 53 \\
			1.6 0 \\
			1.7 0 \\
			1.8 0 \\
			1.9 0 \\
        };

    \addplot+ [ybar interval, opacity=1, color=black, mark=none, fill=gray!60, draw=black]
        table [row sep=\\,x=Bin,y=Equal]{
            Bin Equal \\
            0.0 688 \\
			0.1 2650 \\
			0.2 2157 \\
			0.3 1328 \\
			0.4 775 \\
			0.5 463 \\
			0.6 303 \\
			0.7 179 \\
			0.8 79 \\
			0.9 45 \\
			1.0 29 \\
			1.1 10 \\
			1.2 5 \\
			1.3 4 \\
			1.4 3 \\
			1.5 0 \\
			1.6 0 \\
			1.7 0 \\
			1.8 0 \\
			1.9 0 \\
        };

    \addplot [const plot=false, black, thick, mark=none, domain=0:2, samples=100, smooth]
    {1743.60000 * (
        0.50191 * (1/(sqrt(2*pi*0.16685^2)) * exp(-((x - 1.27087)^2)/(2*0.16685^2)))
      + 0.49809 * (1/(sqrt(2*pi*0.16739^2)) * exp(-((x - 0.28353)^2)/(2*0.16739^2)))
    )};
    
    \addplot+ [ sharp plot, color=blue, thick, mark=none, dashed] coordinates {(1.0229360384733943, 0) (1.0229360384733943, 2651)};

    \addplot+ [ sharp plot, color=olive, thick, mark=none, dashdotdotted] coordinates {(0.8802662875036745, 0) (0.8802662875036745, 2651)};

    \addplot+ [ sharp plot, color=red, thick, mark=none, densely dotted] coordinates {(0.7259594202041626, 0) (0.7259594202041626, 2651)};

    \end{axis}
\end{tikzpicture}}\\ \caption{Plots (a) to (f) show the histogram of pairwise distances computed on the validation set for each dataset. The black line represents the bimodal distribution fitted by our proposed method. Vertical lines show the verification threshold computed using our method (dashed blue) and the \ac{EER} (dotted red). In (g), we show the PKLot 1 histogram for the test set using the threshold computed by \textit{BimodalUpd}.} \label{fig:bimodalsGenerated} \end{figure*}

On average, the proposed \textit{BimodalUpd} method achieves results comparable to the \ac{EER} approach and outperforms the margin-based threshold. 
Unlike the \ac{EER}, however, the proposed method does not require labeled data, enabling the computation of the threshold without any annotation cost.

In Figure \ref{fig:bimodalsGenerated} we present the distance histograms computed for one of the five executions of our experiments, considering the validation set of each dataset. We show the threshold generated by our proposed method and the one generated by the \ac{EER}. We use bars of different styles for positive and negative pairs to illustrate the expected classification errors for each technique. As observed, all datasets exhibit a clear separation between modes, resulting in a pronounced valley between positive and negative distances. 
This explains why different threshold selection strategies yield similar accuracies in balanced scenarios, even when the numerical threshold values differ substantially.

Compared to \textit{BimodalVal}, the \textit{BimodalUpd} configuration consistently benefits from updating the threshold during testing. 
This suggests a mismatch between the validation and test distributions, and indicates that adapting the threshold online can mitigate performance degradation caused by distribution shifts. This effect is illustrated in Figure~\ref{fig:bimodalsGenerated}(g), where the distance distribution observed during testing differs from that of the validation set shown in Figure~\ref{fig:bimodalsGenerated}(d). 
As expected, the test scenario is more challenging, as it involves data from a parking lot unseen during training and validation. 
The green dot-dashed line indicates the final threshold estimated by \textit{BimodalUpd} during test-time adaptation.

\subsection{Non-Normalized Embeddings}\label{sec:non_normalizedTests}

In this experiment, we repeat the evaluation described in Section~\ref{sec:normalizedTest}, removing the L2 normalization layer from the network. 
As a result, the embedding distances become positively unbounded, as in~\cite{sun2014identificationVerification,ribas2024DwellTime}.

\begin{table}[htpb] \caption{Accuracies achieved without L2 normalization.} \label{table:resultsNoL2} \centering \begin{tabular}{lrrrr}\hline & & &\multicolumn{2}{c}{Proposed Methods} \\ & EER & Margin & BimodalVal & BimodalUpd \\\hline MNIST & \textbf{0.99} $\pm$0.00 & 0.50 $\pm$0.00 & \textbf{0.99} $\pm$0.00 & \textbf{0.99} $\pm$0.00 \\ CIFAR-10 & \textbf{0.91} $\pm$0.01 & 0.50 $\pm$0.00 & 0.90 $\pm$0.00 & 0.90 $\pm$0.00 \\ LFW & \textbf{0.88} $\pm$0.01 & 0.50 $\pm$0.00 & 0.86 $\pm$0.01 & 0.86 $\pm$0.00 \\ PKLot 1 & \textbf{0.97} $\pm$0.01 & 0.51 $\pm$0.00 & \textbf{0.97} $\pm$0.00 & 0.96 $\pm$0.01 \\ PKLot 2 & 0.97 $\pm$0.02 & 0.50 $\pm$0.00 & \textbf{0.98} $\pm$0.01 & \textbf{0.98} $\pm$0.01 \\ PKLot 3 & 0.97 $\pm$0.01 & 0.50 $\pm$0.00 & \textbf{0.99} $\pm$0.00 & \textbf{0.99} $\pm$0.00 \\\hline Average & \textbf{0.95} & 0.50 & \textbf{0.95} & \textbf{0.95} \\ \hline \end{tabular} \end{table}

Table~\ref{table:resultsNoL2} summarizes the obtained results. 
Interestingly, with the exception of the margin-based threshold, all methods achieve higher accuracies compared to the normalized embedding setting. 
As in the normalized case, the proposed approach matches the performance of the \ac{EER}-based method, without relying on labeled data.

\begin{figure}[htbp] \centering \begin{tikzpicture}

\begin{axis}[%
hide axis,
xmin=-0.48,
xmax=7.5,
ymin=3.0,
ymax=9.0,
ylabel={$f_2$},
axis x line*=bottom,
axis y line*=left,
legend style={legend cell align=left,align=left,draw=black,legend columns=2, column sep=2.5pt, draw=none,font=\small}
]

\addlegendimage{%
	area legend,
	draw=black,
	fill=gray!80,
	opacity=0.5,
}
\addlegendentry{Positive Pair};

\addlegendimage{%
	area legend,
	draw=black,
	fill=black,
	pattern=north east lines,
	pattern color=black,
	opacity=0.5,
}
\addlegendentry{Negative Pair};

\addlegendimage{color=black,solid,thick}
\addlegendentry{Estimated Bimodal};

\addlegendimage{color=blue,dashed,thick}
\addlegendentry{BimodalVal Threshold};

\addlegendimage{color=red,dotted,thick}
\addlegendentry{EER Threshold};

\end{axis}
\end{tikzpicture}%
 \subfloat[Cifar10]{\hspace{-0.1cm}\begin{tikzpicture}
    \begin{axis}[
        ybar,                                                                       
        width=5.0cm,                                                                
        height=4.8cm,                                                               
        scaled y ticks=base 10:-3,                                                  
        ytick scale label code/.code={},                                            
        yticklabel={\pgfmathprintnumber{\tick}k},                                   
        ticklabel style = {font=\footnotesize},                                     
        label style={font=\footnotesize},                                           
        axis x line=bottom,                                                         
        axis y line=left,           
        xlabel=Distance,
        ylabel=Number of pairs,
        xmin=0, xmax=8.863914489746094
    ]

    \addplot+ [ybar interval, opacity=1, color=gray, pattern = north east lines, pattern color = gray, mark=none]
        table [row sep=\\,x=Bin,y=Diff]{
            Bin Diff \\
            0.0 28 \\
			1.0 6803 \\
			2.0 5247 \\
			3.0 2191 \\
			4.0 1899 \\
			5.0 3597 \\
			6.0 6665 \\
			7.0 3428 \\
        };

    \addplot+ [ybar interval, opacity=1, color=black, mark=none, fill=gray!60, draw=black]
        table [row sep=\\,x=Bin,y=Equal]{
            Bin Equal \\
            0.0 28 \\
			1.0 6779 \\
			2.0 5135 \\
			3.0 1884 \\
			4.0 871 \\
			5.0 279 \\
			6.0 24 \\
			7.0 0 \\
        };

    \addplot [const plot=false, black, thick, mark=none, domain=0:8.863914489746094, samples=100, smooth]
    {30000.00000 * (
        0.47673 * (1/(sqrt(2*pi*0.72053^2)) * exp(-((x - 2.20021)^2)/(2*0.72053^2)))
      + 0.52327 * (1/(sqrt(2*pi*0.93915^2)) * exp(-((x - 6.25645)^2)/(2*0.93915^2)))
    )};
    
    \addplot+ [ sharp plot, color=blue, thick, mark=none, dashed] coordinates {(4.04985348570553, 0) (4.04985348570553, 7918.710076168744)};

    \addplot+ [ sharp plot, color=red, thick, mark=none, densely dotted] coordinates {(4.413238525390625, 0) (4.413238525390625, 7918.710076168744)};

    \end{axis}
\end{tikzpicture}} \subfloat[PKLot1]{\hspace{-0.1cm}\begin{tikzpicture}
    \begin{axis}[
        ybar,                                                                       
        width=5.0cm,                                                                
        height=4.5cm,                                                               
        scaled y ticks=base 10:-3,                                                  
        ytick scale label code/.code={},                                            
        yticklabel={\pgfmathprintnumber{\tick}k},                                   
        ticklabel style = {font=\footnotesize},                                     
        label style={font=\footnotesize},                                           
        axis x line=bottom,                                                         
        axis y line=left,           
        xlabel=Distance,
        xmin=0, xmax=26.76634407043457
    ]

    \addplot+ [ybar interval, opacity=1, color=gray, pattern = north east lines, pattern color = gray, mark=none]
        table [row sep=\\,x=Bin,y=Diff]{
            Bin Diff \\
            0.0 858 \\
			1.0 14265 \\
			2.0 16726 \\
			3.0 11831 \\
			4.0 6063 \\
			5.0 2554 \\
			6.0 1288 \\
			7.0 1374 \\
			8.0 2320 \\
			9.0 3869 \\
			10.0 5617 \\
			11.0 6625 \\
			12.0 6961 \\
			13.0 6424 \\
			14.0 5672 \\
			15.0 4424 \\
			16.0 3203 \\
			17.0 2373 \\
			18.0 1579 \\
			19.0 976 \\
			20.0 644 \\
			21.0 382 \\
			22.0 234 \\
			23.0 118 \\
			24.0 60 \\
			25.0 25 \\
        };

    \addplot+ [ybar interval, opacity=1, color=black, mark=none, fill=gray!60, draw=black]
        table [row sep=\\,x=Bin,y=Equal]{
            Bin Equal \\
            0.0 858 \\
			1.0 14264 \\
			2.0 16724 \\
			3.0 11821 \\
			4.0 6003 \\
			5.0 2377 \\
			6.0 823 \\
			7.0 283 \\
			8.0 65 \\
			9.0 15 \\
			10.0 1 \\
			11.0 0 \\
			12.0 1 \\
			13.0 0 \\
			14.0 0 \\
			15.0 0 \\
			16.0 0 \\
			17.0 0 \\
			18.0 0 \\
			19.0 0 \\
			20.0 0 \\
			21.0 0 \\
			22.0 0 \\
			23.0 0 \\
			24.0 0 \\
			25.0 0 \\
        };

    \addplot [const plot=false, black, thick, mark=none, domain=0:26.76634407043457, samples=100, smooth]
    {106470.00000 * (
        0.47761 * (1/(sqrt(2*pi*1.10120^2)) * exp(-((x - 2.74477)^2)/(2*1.10120^2)))
      + 0.52239 * (1/(sqrt(2*pi*3.45662^2)) * exp(-((x - 12.87438)^2)/(2*3.45662^2)))
    )};
    
    \addplot+ [ sharp plot, color=blue, thick, mark=none, dashed] coordinates {(6.071207343296453, 0) (6.071207343296453, 18422.345478005518)};

    \addplot+ [ sharp plot, color=red, thick, mark=none, densely dotted] coordinates {(6.691476345062256, 0) (6.691476345062256, 18422.345478005518)};

    \end{axis}
\end{tikzpicture}}\\ \caption{Histogram of pairwise distances computed on the validation sets considering embeddings without L2 normalization.} \label{fig:histoNoL2} \end{figure}
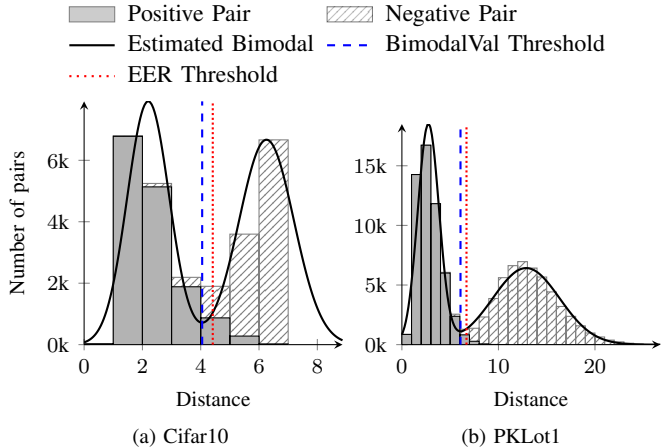

The poor performance of the margin-based threshold in this scenario highlights its unsuitability when distances are unbounded. 
Despite this, the distance distributions remain clearly bimodal, and the mode corresponding to negative pairs does not exceed 15 units in any dataset. 
This behavior is illustrated in Figure~\ref{fig:histoNoL2}, and was consistently observed across all evaluated datasets.

\subsection{Unbalanced Scenario}\label{sec:unbalancedTest}

We now evaluate the proposed approach under unbalanced testing conditions, where positive pairs are significantly more frequent than negative ones. 
Such scenarios commonly arise in real-world applications. 
For example, in dwell time estimation using the PKLot dataset~\cite{ribas2024DwellTime}, vehicles may be compared sequentially over time, resulting in a high proportion of positive~matches.

To simulate this setting, the training procedure remains unchanged. 
During testing, however, each sample generates a positive pair with 90\% probability and a negative pair with 10\% probability, yielding a 9:1 imbalance. 
Due to this imbalance, performance is reported using balanced accuracy.

\begin{table}[htpb] \caption{Balanced accuracies achieved considering the tested methods under unbalanced scenarios.} \label{table:resultsUnbal} \centering \begin{tabular}{lrrrr}\hline & & &\multicolumn{2}{c}{Proposed Methods} \\ & EER & Margin & BimodalVal & BimodalUpd \\\hline MNIST & \textbf{1.00} $\pm$0.00 & \textbf{1.00} $\pm$0.00 & \textbf{1.00} $\pm$0.00 & \textbf{1.00} $\pm$0.00 \\ CIFAR-10 & \textbf{1.00} $\pm$0.00 & \textbf{1.00} $\pm$0.00 & 0.96 $\pm$0.04 & 0.93 $\pm$0.03 \\ LFW & 0.78 $\pm$0.05 & 0.76 $\pm$0.05 & 0.85 $\pm$0.04 & \textbf{0.87} $\pm$0.03 \\ PKLot 1 & 0.97 $\pm$0.05 & 0.92 $\pm$0.10 & 0.92 $\pm$0.10 & \textbf{1.00} $\pm$0.00 \\ PKLot 2 & 0.97 $\pm$0.05 & 0.95 $\pm$0.05 & 0.95 $\pm$0.05 & \textbf{0.99} $\pm$0.00 \\ PKLot 3 & 0.99 $\pm$0.00 & \textbf{1.00} $\pm$0.00 & \textbf{1.00} $\pm$0.00 & 0.98 $\pm$0.01 \\\hline Average & 0.95 & 0.94 & 0.95 & \textbf{0.96} \\\hline \end{tabular} \end{table}

As shown in Table~\ref{table:resultsUnbal}, the \textit{BimodalUpd} method achieves the highest balanced accuracy across datasets, demonstrating its ability to adapt the verification threshold even under severe class imbalance.

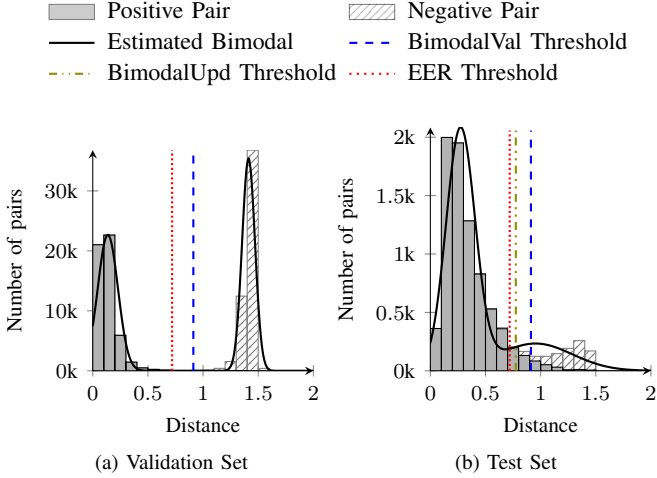
\begin{figure}[htbp] \centering \begin{tikzpicture}

\begin{axis}[%
hide axis,
xmin=-0.48,
xmax=7.5,
ymin=3.0,
ymax=9.0,
ylabel={$f_2$},
axis x line*=bottom,
axis y line*=left,
legend style={legend cell align=left,align=left,draw=black,legend columns=2, column sep=2.5pt, draw=none,font=\small}
]

\addlegendimage{%
	area legend,
	draw=black,
	fill=gray!80,
	opacity=0.5,
}
\addlegendentry{Positive Pair};

\addlegendimage{%
	area legend,
	draw=black,
	fill=black,
	pattern=north east lines,
	pattern color=black,
	opacity=0.5,
}
\addlegendentry{Negative Pair};

\addlegendimage{color=black,solid,thick}
\addlegendentry{Estimated Bimodal};

\addlegendimage{color=blue,dashed,thick}
\addlegendentry{BimodalVal Threshold};

\addlegendimage{color=olive,dashdotdotted,thick}
\addlegendentry{BimodalUpd Threshold};

\addlegendimage{color=red,dotted,thick}
\addlegendentry{EER Threshold};

\end{axis}
\end{tikzpicture}%
 \subfloat[Validation Set]{\hspace{-0.1cm}\begin{tikzpicture}
    \begin{axis}[
        ybar,                                                                       
        width=4.5cm,                                                                
        height=4.5cm,                                                               
        scaled y ticks=base 10:-3,                                                  
        ytick scale label code/.code={},                                            
        yticklabel={\pgfmathprintnumber{\tick}k},                                   
        ticklabel style = {font=\footnotesize},                                     
        label style={font=\footnotesize},                                           
        axis x line=bottom,                                                         
        axis y line=left,           
        xlabel=Distance,
        ylabel=Number of pairs,
        xmin=0, xmax=2
    ]

    \addplot+ [ybar interval, opacity=1, color=gray, pattern = north east lines, pattern color = gray, mark=none]
        table [row sep=\\,x=Bin,y=Diff]{
            Bin Diff \\
            0.0 21027 \\
			0.1 22645 \\
			0.2 5925 \\
			0.3 1449 \\
			0.4 510 \\
			0.5 258 \\
			0.6 130 \\
			0.7 65 \\
			0.8 66 \\
			0.9 99 \\
			1.0 150 \\
			1.1 406 \\
			1.2 1541 \\
			1.3 12450 \\
			1.4 36793 \\
			1.5 399 \\
			1.6 2 \\
			1.7 0 \\
			1.8 0 \\
			1.9 0 \\
        };

    \addplot+ [ybar interval, opacity=1, color=black, mark=none, fill=gray!60, draw=black]
        table [row sep=\\,x=Bin,y=Equal]{
            Bin Equal \\
            0.0 21024 \\
			0.1 22629 \\
			0.2 5912 \\
			0.3 1435 \\
			0.4 481 \\
			0.5 226 \\
			0.6 108 \\
			0.7 30 \\
			0.8 17 \\
			0.9 38 \\
			1.0 23 \\
			1.1 20 \\
			1.2 14 \\
			1.3 1 \\
			1.4 0 \\
			1.5 0 \\
			1.6 0 \\
			1.7 0 \\
			1.8 0 \\
			1.9 0 \\
        };

    \addplot [const plot=false, black, thick, mark=none, domain=0:2, samples=100, smooth]
    {10391.50000 * (
        0.50122 * (1/(sqrt(2*pi*0.09164^2)) * exp(-((x - 0.13672)^2)/(2*0.09164^2)))
      + 0.49878 * (1/(sqrt(2*pi*0.05840^2)) * exp(-((x - 1.41216)^2)/(2*0.05840^2)))
    )};
    
    \addplot+ [ sharp plot, color=blue, thick, mark=none, dashed] coordinates {(0.9118855258464718, 0) (0.9118855258464718, 36793)};

    \addplot+ [ sharp plot, color=red, thick, mark=none, densely dotted] coordinates {(0.7193761467933655, 0) (0.7193761467933655, 36793)};

    \end{axis}
\end{tikzpicture}} \subfloat[Test Set]{\hspace{-0.1cm}\begin{tikzpicture}
    \begin{axis}[
        ybar,                                                                       
        width=4.5cm,                                                                
        height=4.8cm,                                                               
        scaled y ticks=base 10:-3,                                                  
        ytick scale label code/.code={},                                            
        yticklabel={\pgfmathprintnumber{\tick}k},                                   
        ticklabel style = {font=\footnotesize},                                     
        label style={font=\footnotesize},                                           
        axis x line=bottom,                                                         
        axis y line=left,           
        xlabel=Distance,
        ylabel=Number of pairs,
        xmin=0, xmax=2
    ]

    \addplot+ [ybar interval, opacity=1, color=gray, pattern = north east lines, pattern color = gray, mark=none]
        table [row sep=\\,x=Bin,y=Diff]{
            Bin Diff \\
            0.0 362 \\
			0.1 1998 \\
			0.2 1951 \\
			0.3 1285 \\
			0.4 832 \\
			0.5 532 \\
			0.6 369 \\
			0.7 205 \\
			0.8 166 \\
			0.9 124 \\
			1.0 124 \\
			1.1 147 \\
			1.2 192 \\
			1.3 258 \\
			1.4 170 \\
			1.5 3 \\
			1.6 0 \\
			1.7 0 \\
			1.8 0 \\
			1.9 0 \\
        };

    \addplot+ [ybar interval, opacity=1, color=black, mark=none, fill=gray!60, draw=black]
        table [row sep=\\,x=Bin,y=Equal]{
            Bin Equal \\
            0.0 362 \\
			0.1 1998 \\
			0.2 1951 \\
			0.3 1284 \\
			0.4 828 \\
			0.5 529 \\
			0.6 360 \\
			0.7 196 \\
			0.8 131 \\
			0.9 84 \\
			1.0 52 \\
			1.1 31 \\
			1.2 9 \\
			1.3 11 \\
			1.4 6 \\
			1.5 0 \\
			1.6 0 \\
			1.7 0 \\
			1.8 0 \\
			1.9 0 \\
        };

    \addplot [const plot=false, black, thick, mark=none, domain=0:2, samples=100, smooth]
    {871.80000 * (
        0.77611 * (1/(sqrt(2*pi*0.13122^2)) * exp(-((x - 0.27342)^2)/(2*0.13122^2)))
      + 0.22389 * (1/(sqrt(2*pi*0.33363^2)) * exp(-((x - 0.95330)^2)/(2*0.33363^2)))
    )};
    
    \addplot+ [ sharp plot, color=blue, thick, mark=none, dashed] coordinates {(0.9118855258464718, 0) (0.9118855258464718, 2057.032827328811)};

    \addplot+ [ sharp plot, color=olive, thick, mark=none, dashdotdotted] coordinates {(0.774753, 0) (0.774753, 2057.032827328811)};

    \addplot+ [ sharp plot, color=red, thick, mark=none, densely dotted] coordinates {(0.7193761467933655, 0) (0.7193761467933655, 2057.032827328811)};

    \end{axis}
\end{tikzpicture}}\\ \caption{Distance histograms and the computed thresholds for the unbalanced test scenario considering the PKLot 1 test scenario.} \label{fig:unbalancedScenario} \end{figure}

Figure~\ref{fig:unbalancedScenario} presents the distance distributions for the validation and test sets in the PKLot 1 scenario. 
Although the proposed method successfully identifies an appropriate threshold in this case, we acknowledge that extreme imbalance may pose challenges, as one of the modes may become difficult to identify. 
In such situations, the optimization procedure may converge to a minimum outside the interval between the true modes. 
Nevertheless, this behavior was not observed in our experiments.

\section{Conclusions}\label{sec:conclusion}

In this work, we proposed a method to estimate the verification threshold for siamese verification networks based on the assumption that the distribution of distances produced by such networks can be approximated by a bimodal function. The proposed approach was evaluated using four well-known datasets covering distinct verification tasks: digit comparison for MNIST, object classification across 10 classes for CIFAR-10, face verification for LFW, and car verification for PKLot.

The experimental results show that the proposed method consistently outperforms the direct use of the margin value as a verification threshold and achieves performance comparable to thresholds computed using the \ac{EER}. Our proposed approach does not require labeled data, which represents a practical advantage over traditional validation-based threshold selection strategies. Moreover, the method enables the verification threshold to be updated during deployment as new data becomes available, also without supervision. Our experiments indicate that such online updates can be beneficial in scenarios where the test distribution differs from the training or validation distributions.

We further evaluated the proposed approach under unbalanced testing conditions, which are common in real-world verification systems. The results show that the method is able to estimate a reasonable classification threshold even in the presence of class imbalance. Nevertheless, under more extreme imbalance ratios, the method may fail to identify one of the two modes of the distance distribution, potentially leading to suboptimal threshold estimation.

As future research, we also intend to study mechanisms to improve the robustness of the proposed method under severe imbalance and distribution shifts, including alternative distribution modeling strategies and adaptive windowing schemes. Additionally, we plan to further analyze the behavior of online threshold updating under dynamic environments.

\bibliographystyle{IEEEtran}
\bibliography{bibliography}

\end{document}